\pdfoutput=1

\documentclass[11pt]{article}

\usepackage[preprint]{acl}
\usepackage{comment}
\usepackage{times}
\usepackage{colortbl}

\usepackage{latexsym}
\usepackage{booktabs}

\usepackage[T1]{fontenc}

\usepackage[utf8]{inputenc}

\usepackage{microtype}
\usepackage{adjustbox}
\usepackage{inconsolata}
\usepackage{tabularx}
\usepackage{graphicx}
\usepackage{amsmath}
\usepackage{amssymb}
\usepackage{multicol}
\usepackage{multirow}
\NewDocumentCommand{\jiayu}
{ mO{} }{\textcolor{blue}{\textsuperscript{\textit{jiayu}}\textsf{\textbf{\small[#1]}}}}

\title{Revisiting Epistemic Markers in Confidence Estimation: Can Markers Accurately Reflect Large Language Models' Uncertainty?}

\author{Jiayu Liu,
Qing Zong,
Weiqi Wang,
Yangqiu Song\\
Department of Computer Science and Engineering, HKUST, Hong Kong SAR, China\\
\texttt{jliufv@connect.ust.hk; \{qzong, wwangbw, yqsong\}@cse.ust.hk}\\
}

\begin{document}
\maketitle

\begin{abstract}
As Large Language Models (LLMs) are increasingly used in high-stakes domains, accurately assessing their confidence is crucial. 
Humans typically express confidence through epistemic markers (e.g., ``fairly confident'') instead of numerical values.
However, it remains unclear whether LLMs reliably use these markers to reflect their intrinsic confidence due to the difficulty of quantifying uncertainty associated with various markers.
To address this gap, we first define \emph{marker confidence} as the observed accuracy when a model employs an epistemic marker. 
We evaluate its stability across multiple question-answering datasets in both in-distribution and out-of-distribution settings for open-source and proprietary LLMs. 
Our results show that while markers generalize well within the same distribution, their confidence is inconsistent in out-of-distribution scenarios. 
These findings raise significant concerns about the reliability of epistemic markers for confidence estimation, underscoring the need for improved alignment between marker based confidence and actual model uncertainty.
Our code is available at \url{https://github.com/HKUST-KnowComp/MarConf}.
\end{abstract}

\section{Introduction}

\begin{figure}[t]
     \centering
     \includegraphics[width=1\linewidth]{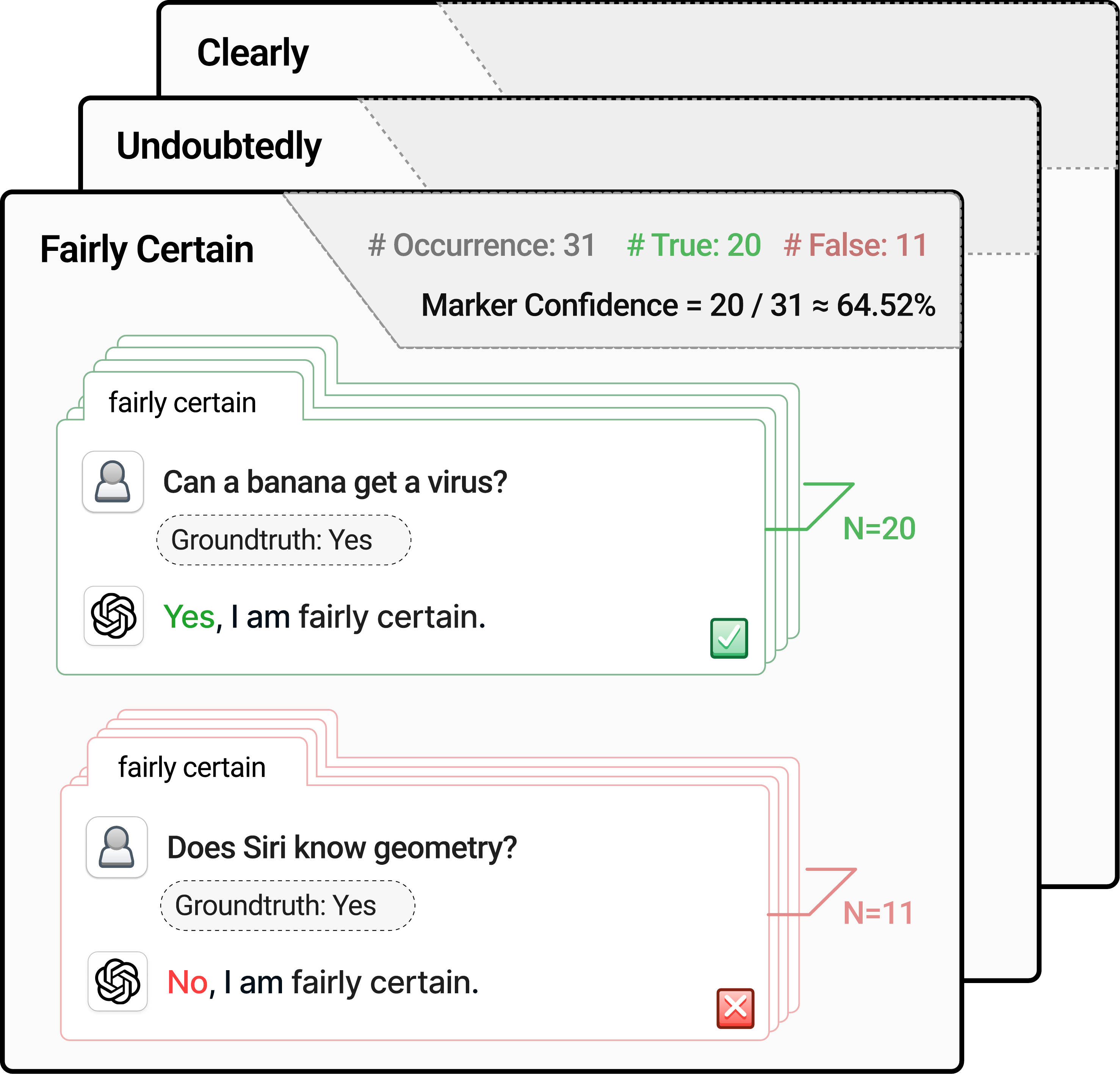}
     \vspace{-0.1in}
     \caption{An example of our framework calculating the marker confidence of ``fairly certain'' for \textit{GPT-4o} on StrategyQA. We calculate the confidence for all the markers across seven models and seven datasets.}
    \label{fig:marker-confidence-pipeline}
    \vspace{-0.2in}
\end{figure}

LLMs have grown increasingly powerful, yet their application in mission-critical tasks is still hindered by reliability issues \cite{LLM-not-reliable, hallucination-original}.
Therefore, accurately measuring output confidence is crucial for their reliable deployment \cite{confidence-extraction-important-1, confidence-extraction-important-2, confidence-estimation-important-3}.
Traditionally, black-box confidence estimation in LLMs primarily relies on direct numerical outputs (e.g., “30\% confidence”) or response consistency \cite{xiongcan, BSDetector, thinktwicebeforetrust}, while white-box methods mainly utilize logits, internal states as information source \cite{confidenceestimationsurvey}.
However, natural language is the primary interface for human-LLM interaction. 
Instead of relying solely on abstract numerical measures, humans often use epistemic markers, such as ``I am not sure'' or ``it is unlikely that,'' to convey uncertainty \cite{valuemeaningofmarker1, MarkerVSNumber, vaguemeaningofmarker2, LLMknowmostlywhattheyknow}. 
This similar recognition of uncertainty markers is essential for effective communication \cite{nodblp-HumanAlignmengInProbabilityTerm,PerceptionOfLinguisticUncertainty}, which potentially makes it valuable for LLMs to adopt a similar practice \cite{intrinsicuncertaintyinwords}. 

However, it remains unexplored whether LLMs are capable of incorporating these epistemic markers in their responses to express their intrinsic confidence stably and consistently.
Previous works have primarily concentrated on the misalignment between human and LLM recognition of epistemic markers \cite{zhou2024relying, tangzhishengnaive, PerceptionOfLinguisticUncertainty}, concluding that models always fail to accurately convey confidence in words \cite{intrinsicuncertaintyinwords}.
In fact, human interpretations of markers are not completely identical \cite{confidence-is-accurate-in-crowd-source-scenarios(explaining-10)-and-people-intepret-differently-to-markers}, so even if these markers may not align well with human reasoning, they can still be useful if the model maintains a consistent internal mapping between markers and their actual accuracy. 
Thus, previous studies questioning the reliability of markers may be insufficient, as they have not examined whether LLMs can consistently apply their own confidence framework.

To address this gap, we investigate whether epistemic markers produced by LLMs reliably reflect their confidence in question-answering tasks. 
By defining marker confidence as the accuracy of responses when a model uses a specific marker to convey confidence, we calculate the marker confidence with various models and datasets. Additionally, we propose seven evaluation metrics to systematically assess the stability of these markers in both in-distribution and out-of-distribution contexts.
Our findings show that while markers perform well within similar distributions, their stability declines in out-of-distribution contexts. 
Additionally, we compare a range of widely used models and conclude that the more powerful ones demonstrate a better understanding of epistemic markers.

\section{Related Work}

Our work primarily intersects with confidence estimation in LLMs and studies about epistemic markers.
Please find related works in Appendix~\ref{app:related-work}.

\section{Study Design}

\subsection{Formalization}

\textbf{Confidence of Epistemic Markers.} Let \(W\) denote an epistemic marker, \(D = \{Q_1, Q_2, \dots, Q_n\}\) a labeled dataset, and \(M\) a model. We define the confidence associated with each epistemic marker as \(\textit{Conf}(W, D, M)\), computed as the accuracy of the answers that explicitly include marker \(W\) when the model provides responses. 
It is important to note that our definition of marker confidence deviates from the conventional interpretation of verbal confidence, which pertains to the semantic uncertainty associated with epistemic markers.
To compute the marker confidence for a specific epistemic marker \(W_i\), we use model \(M_k\) to generate answers for all questions in the training set of dataset \(D_j\) and then extract the subset \(Q_{W_i} \subseteq D_j\) consisting of questions whose generated answers contain \(W_i\). The marker confidence is defined as:

{\small
\begin{align*}
\textit{Conf}(W_i, D_j, M_k) = 
\frac{1}{\left| Q_{W_i} \right|} 
\sum_{q \in Q_{W_i}} 
\mathbb{I} \Big(M_k(q) \Big),
\end{align*}}

\noindent where $Q_{W_i}$ is the set of questions whose generated answers contain the epistemic marker $W_i$ and $\mathbb{I}(\cdot)$ is the indicator function, which is 1 if the answer generated by $M_k$ for question $q$ is correct, and 0 otherwise. An example is provided in \autoref{fig:marker-confidence-pipeline}.

\begin{table*}[h]
\small
\centering
\resizebox{\linewidth}{!}{%
\begin{tabular}{l c c c c c c c}
\toprule
 \multirow{2}{*}{\textbf{Model}} & \multicolumn{5}{c}{\textbf{Marker Confidence}} & \multicolumn{1}{c}{\textbf{Rank}} & \multicolumn{1}{c}{\textbf{Density}} \\
\cmidrule(lr){2-6} \cmidrule(lr){7-7} \cmidrule(lr){8-8}
 & I-AvgECE $\downarrow$ & C-AvgECE $\downarrow$ & NumECE $\downarrow$ & C-AvgCV $\downarrow$ & MAC & MRC $\uparrow$ & I-AvgCV \\
\midrule
\textit{Llama-3.1-8B-Instruct}      & 10.09 & 15.95 & 22.70 & 20.80 & 60.91 & 11.37 & 20.48 \\
\textit{Qwen2.5-7B-Instruct}        & \phantom{0}7.85 & 23.60 & 21.84 & 31.29 & 68.06 & 11.85 & 22.39 \\
\textit{Qwen2.5-14B-Instruct}       & \phantom{0}7.66 & 20.38 & 17.98 & 26.44 & 73.95 & 34.60 & 23.83 \\
\textit{Qwen2.5-32B-Instruct}       & \textbf{\phantom{0}4.78} & 10.40 & \phantom{0}8.86 & 19.24 & 78.20 & \textbf{36.97} & 16.26 \\
\textit{Mistral-7B-Instruct-v0.3}   & 10.58 & 24.81 & 24.46 & 28.52 & 84.57 & 10.54 & 21.01 \\
\textit{GPT-4o}                     & \phantom{0}8.55 & \textbf{11.84} & \textbf{\phantom{0}7.56} & \textbf{15.72} & 76.44 & 27.54 & 14.30 \\
\textit{GPT-4o-mini}                & \phantom{0}7.65 & 17.15 & 12.79 & 21.98 & 87.68 & 16.48 & 20.61 \\
\midrule
Average                    & \phantom{0}8.17 & 17.73 & 16.60 & 23.43 & 75.69 & 21.34 & 19.84 \\
\bottomrule
\end{tabular}%
}
\caption{Model performance across seven metrics. For each metric, the data for the best performing model is bolded. For analytical experiments about markers, we only consider those appear no less than 10 times to eliminate the effect of randomness (see Section~\ref{why marker > 10} for details). All values listed in the table are expressed as percentage (\%).}
\label{tab:combined_metrics}
\vspace{-0.1in}
\end{table*}

\subsection{Methods}

We calculate \textit{Conf}($W_i, D_j, M_k$) for all combinations of generated markers, datasets and models ($W_i, D_j, M_k$) in Appendix~\ref{app:experiment-setup} to provide an all-rounded insight into the marker distributions. 
Specifically, we propose seven metrics to systematically evaluate the stability and consistency of LLM generated epistemic markers:

\noindent\textbf{(1) I-AvgECE} \textit{In-domain Average ECE} reflects how well the marker confidence of the model aligns with its actual accuracy in a consistent setting within the same distribution.

\noindent\textbf{(2) C-AvgECE} \textit{Cross-domain Average ECE} assesses the calibration error of the marker confidence and the actual accuracy, further reflecting the robustness of the model’s marker confidence in out-of-domain scenarios.

\noindent\textbf{(3) NumECE} \textit{Numerical ECE} measures the overall calibration of the model’s numerical confidence outputs across all datasets. 
All ECE-related metrics are desired with a lower value, indicating better calibration performance on the target dataset.

\noindent\textbf{(4) MAC} \textit{Marker Accuracy Correlation} reflects the correlation between marker confidence and the model accuracy on different datasets. The metric is calculated based on Pearson coefficient, so 0 in this value represents no linear correlation and 1 indicates direct proportional relationship between the marker confidence and model's accuracy.

\noindent\textbf{(5) MRC} \textit{Marker Ranking Correlation} measures the model's ability to maintain a consistent marker confidence ranking across different datasets.
The metric is calculated based on Spearman coefficient \cite{coefficients}, so 0 in this value represents no correlation and 1 indicates totally identical between markers' confidence rankings.

\noindent\textbf{(6) I-AvgCV} \textit{In-domain Average CV} captures the dispersion of the model-generated confidence scores within each dataset. A relatively higher I-AvgCV value indicates a more decentralized distribution of markers within the dataset, demonstrating a stronger ability to distinguish between different markers.

\noindent\textbf{(7) C-AvgCV} \textit{Cross-domain Average CV} measures the consistency of the model’s marker-based confidence across different datasets. A higher C-AvgCV value indicates a greater dispersion of marker confidence across various datasets, suggesting the model’s instability regarding marker confidence.

More details about the design and implementation of the metrics can be found in Appendix~\ref{app:evaluation-metrics-appendix}.

\section{Experiments and Analysis}


\paragraph{Models and Datasets.}
We experiment with two mainstream open-source and five proprietary LLMs over seven datasets from various domains.
More introduction can be found in Appendix~\ref{app:experiment-setup}. 

\paragraph{Baseline.} 
Inspired by previous comparison about using numerical values and uncertainty expression in words to express confidence level \cite{number-marker-comparison-1, number-marker-comparison-2}, we apply the method of directly prompting the model to express a numerical confidence as baseline for comparison. The prompt designed to elicit epistemic markers and numerical confidence is in Appendix~\ref{app:prompt}.
Our main experiment results are in Table~\ref{tab:combined_metrics}. 

\paragraph{Marker Filtering Strategies}
\label{why marker > 10}

We conduct all marker analysis experiments (namely C-AvgCV, MAC, MRC, and I-AvgCV) by filtering markers that occur fewer than 10 times in the training set in Table~\ref{tab:combined_metrics}. 
The filtering threshold is eventually a tradeoff between the completeness and reliability of the data. 
On one hand, if the filtering threshold (10 in the main table) is too small, the confidence interval of the results would be large. On the other hand, we cannot include diverse markers for the metrics and get limited insights. 
The confidence intervals and more details are reported in Appendix~\ref{app:why-filtering}.

\begin{figure*}[h]
\centering
    \begin{minipage}{0.5\textwidth}
        \centering
        \includegraphics[width=\textwidth]{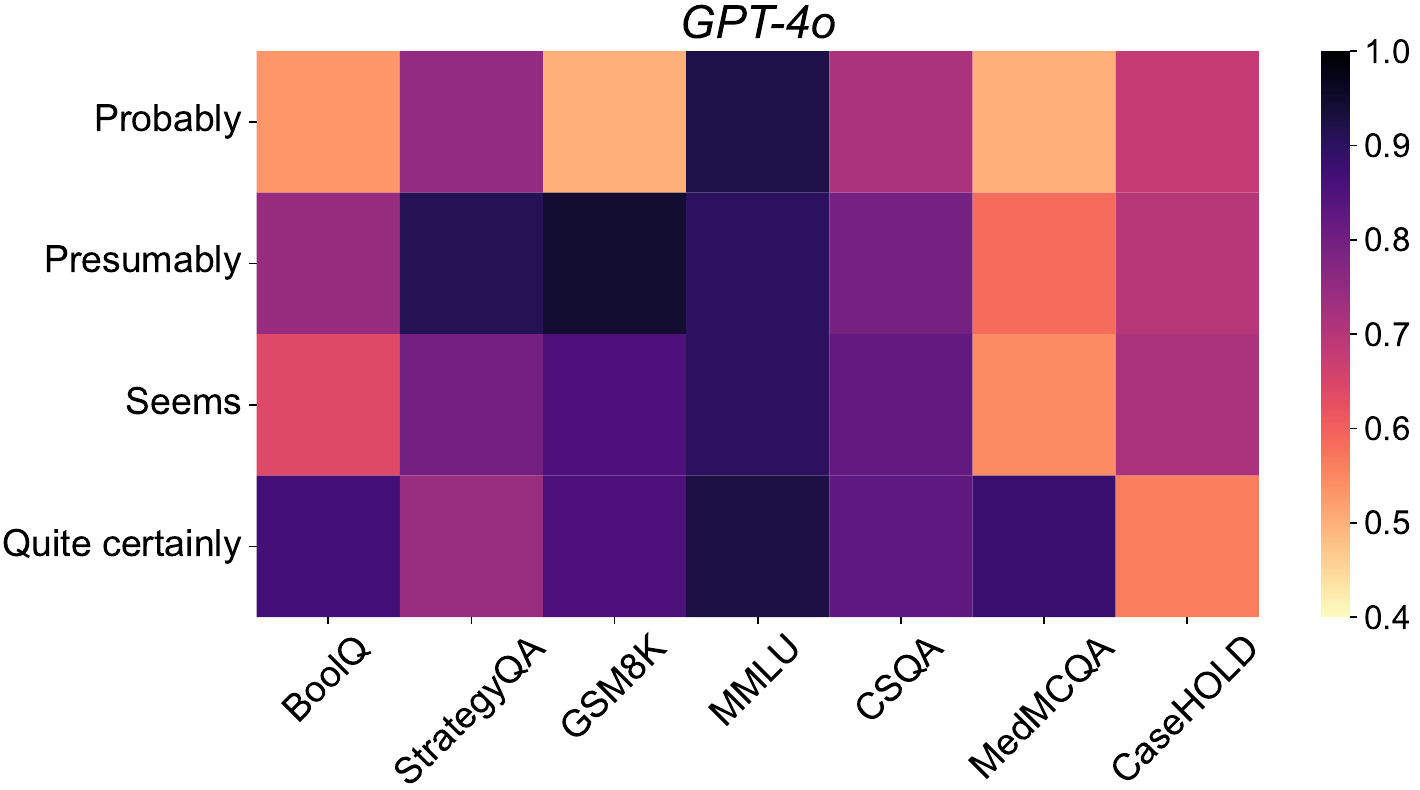}
    \end{minipage}%
    \hfill
    \begin{minipage}{0.5\textwidth}
        \centering
        \includegraphics[width=\textwidth]{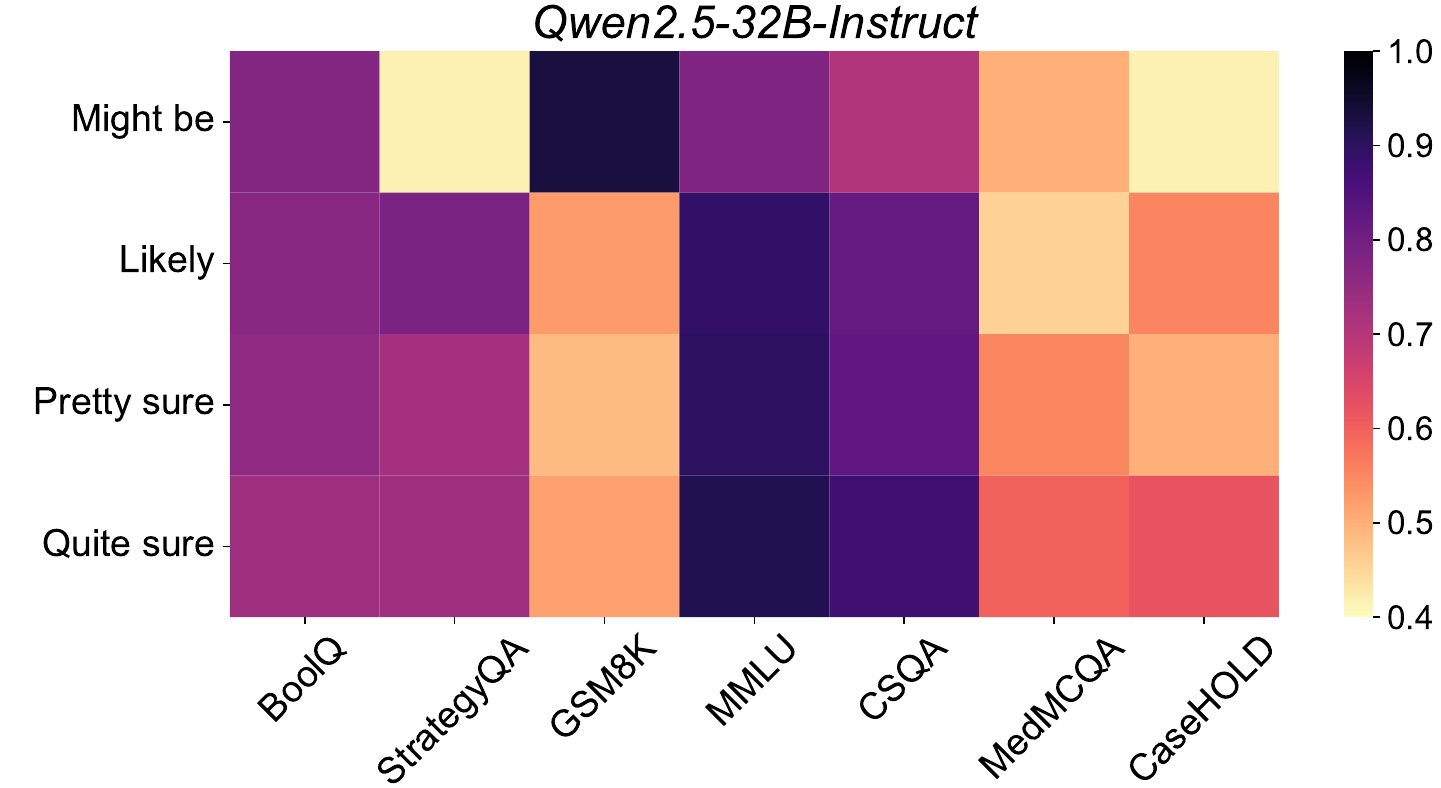}
    \end{minipage}
    \caption{\textbf{Model's marker confidence varies greatly across different datasets.} 
    We plot the heatmap of the marker confidence of \textit{GPT-4o} and \textit{Qwen2.5-32B-Instruct} across different datasets, illustrating that even the best models exhibit substantially different confidence levels in various contexts. The markers in the graph are randomly selected from the shared markers of all datasets.}
    \label{fig:heat-map-marker-confidence}
    \vspace{-0.1in}
\end{figure*}

\begin{figure*}[h]
\centering
    \begin{minipage}{0.5\textwidth}
        \centering
        \includegraphics[width=\textwidth]{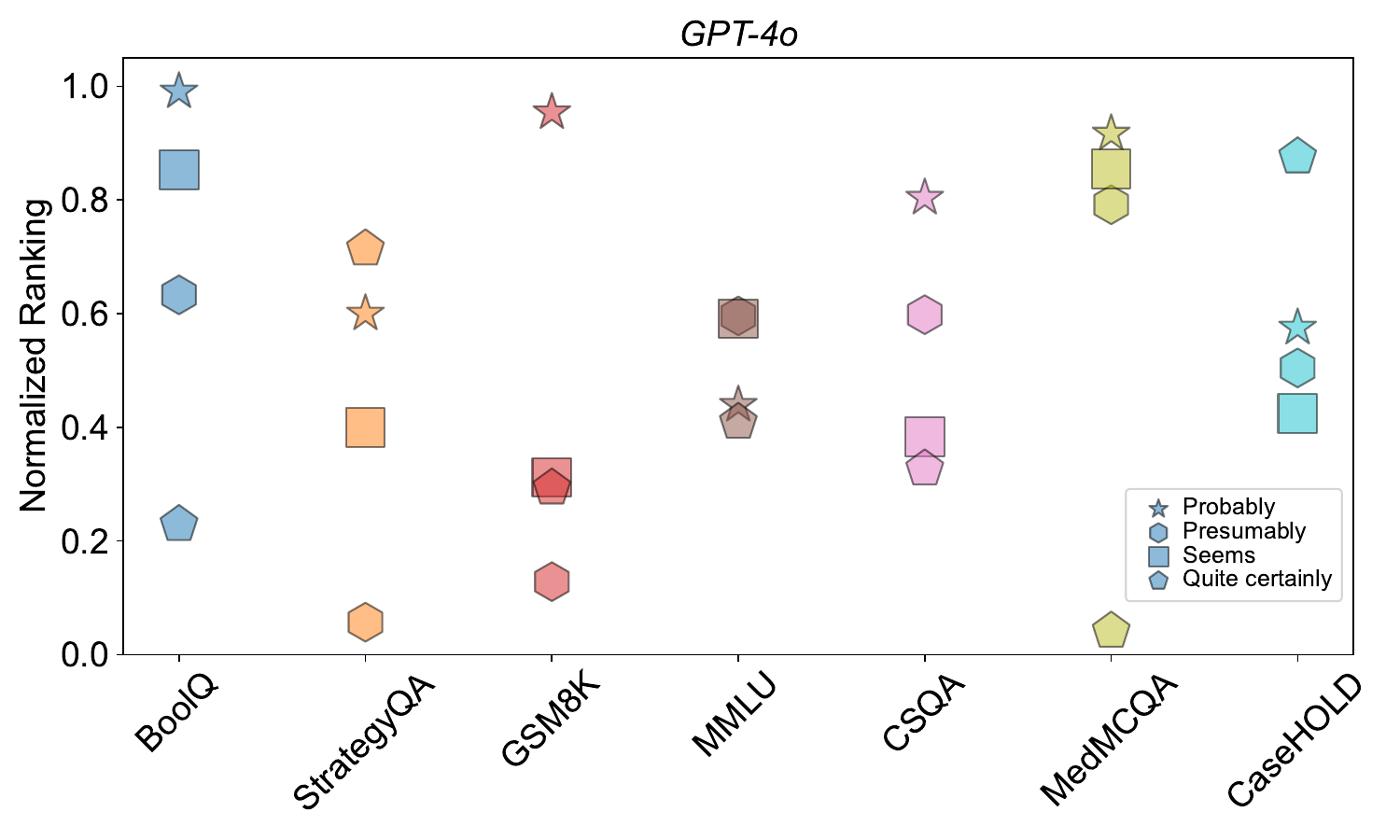}
    \end{minipage}%
    \hfill
    \begin{minipage}{0.5\textwidth}
        \centering
        \includegraphics[width=\textwidth]{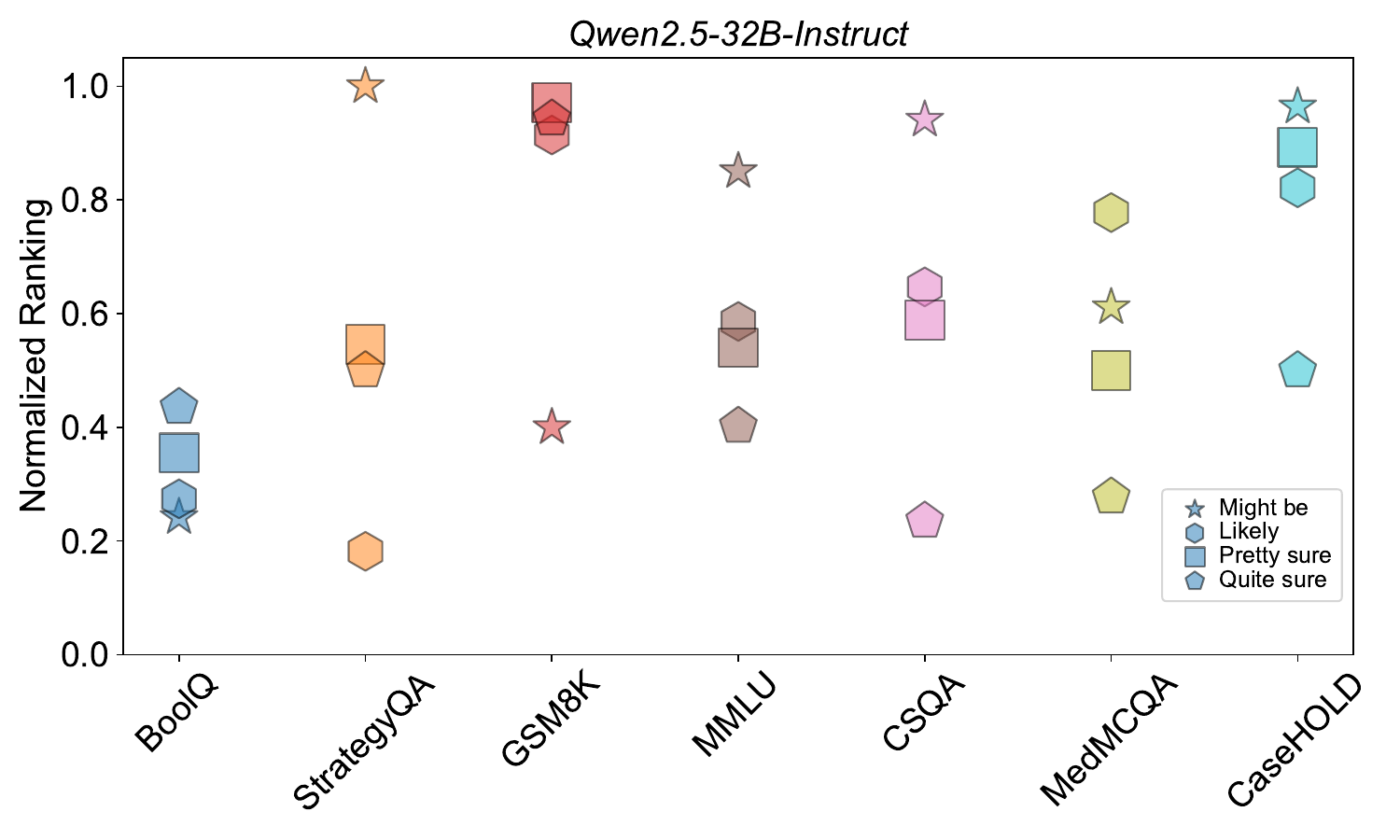}
    \end{minipage}
    \caption{\textbf{The rankings of the model's marker confidence fluctuates significantly across different datasets.} We plot the scatter diagram for the marker confidence rankings of the best performing models, but still discovered that the rankings are extremely unstable. The markers in the graph are randomly selected from the shared markers across all datasets.}
    \label{fig:plot-map-marker-ranking}
    \vspace{-0.15in}
\end{figure*}

\subsection{Main Observations}

\label{subsec:findings}

\paragraph{While the in-distribution marker confidence is relatively stable, it lacks robustness across different datasets.} This conclusion is supported by the observation that I-AvgECE values consistently remain lower than NumECE and C-AvgECE values are notably higher than NumECE for 6 out of 7 models, indicating that models exhibit shortcomings in generalizing marker confidence to different distributions and datasets.

More direct evidence to support the conclusion may be inferred from the C-AvgCV of the models.
The average C-AvgCV of 0.2343 highlights that marker confidence is highly sensitive to distribution shifts, aligning with the observed high value in C-AvgECE. 
Notably, we observed that stronger models (e.g., \textit{GPT-4o, Qwen2.5-32B-Instruct}) might exhibit smaller C-AvgCV values.

We further examine the relationship between marker confidence and model accuracy across different datasets using MAC. 
For 5 out of 7 of the models, the MAC value is over 0.7, which indicates that marker confidence are positively related to the model's accuracy in a strict manner. 
This also suggests that marker confidence is fragile under distribution shifts, highlighting models' lack of robust understanding of epistemic markers.

\paragraph{Models fail to maintain a consistent ordering of epistemic markers across different domains.} The overall low values of MRC suggest that models do not preserve a consistent ranking of markers when applied to datasets with different distributions. 
We notice that larger models appear to have a better grasp at maintaining a stable ordering of markers. However, both the maximum and average MRC indicates low consistency performance, suggesting a lack of robust understanding of marker relative confidence.

\begin{table*}[h!]
  \centering
  \small
  \begin{tabular}{p{3.25cm} >{\centering\arraybackslash}p{1.1cm} >{\centering\arraybackslash}p{1.1cm} >{\centering\arraybackslash}p{1.1cm} >{\centering\arraybackslash}p{1.1cm} >{\centering\arraybackslash}p{1.1cm} >{\centering\arraybackslash}p{1.1cm} >{\centering\arraybackslash}p{1.1cm} >{\centering\arraybackslash}p{1.1cm}}
    \toprule
    \textbf{Model} & \multicolumn{2}{c}{\textbf{C-AvgCV $\downarrow$}} & \multicolumn{2}{c}{\textbf{MAC}} & \multicolumn{2}{c}{\textbf{MRC $\uparrow$}} & \multicolumn{2}{c}{\textbf{I-AvgCV}} \\
    \cmidrule(lr){1-1} \cmidrule(lr){2-3} \cmidrule(lr){4-5} \cmidrule(lr){6-7} \cmidrule(lr){8-9} 
    \textbf{Threshold} & \textbf{50} & \textbf{100} & \textbf{50} & \textbf{100} & \textbf{50} & \textbf{100} & \textbf{50} & \textbf{100} \\
    \midrule

    \textit{Llama-3.1-8B-Instruct} & 25.50 & 23.39 & 77.54 & 65.98 & -8.57 & -10.32 & 11.44 & \phantom{0}8.59 \\
    \textit{Qwen2.5-7B-Instruct} & 30.12 & 29.30 & 77.96 & 69.03 & \phantom{0}6.01 & \phantom{0}2.25 & 18.38 & 15.41 \\
    \textit{Qwen2.5-14B-Instruct} & 26.49 & 27.66 & 92.03 & 94.25 & 36.56 & 36.11 & 20.59 & 21.13 \\
    Qwen2.5-32B-Instruct & 20.03 & 21.27 & 87.25 & 86.85 & 34.91 & 23.95 & 13.51 & 12.18 \\
    \textit{Mistral-7B-Instruct-v0.3} & 26.70 & 26.34 & 94.80 & 84.71 & 30.68 & 30.60 & 12.81 & \phantom{0}9.67 \\
    \textit{GPT-4o} & \textbf{15.86} & \textbf{16.92} & 89.91 & 90.68 & \textbf{37.38} & \textbf{39.40} & \phantom{0}7.52 & \phantom{0}7.29 \\
    \textit{GPT-4o-mini} & 22.15 & 22.41 & 86.82 & 87.45 & 24.16 & 24.36 & 11.98 & 11.39 \\
    \midrule
    Average & 23.84 & 23.90 & 86.62 & 82.71 & 23.02 & 20.91 & 13.75 & 12.24 \\
    \bottomrule
  \end{tabular}
  \caption{This table presents the results (in \%) of the marker analysis experiments, organized by different filtering thresholds. We observed that the conclusion obtained from Table~\ref{tab:combined_metrics} still holds when the threshold increases. }
  \label{tab:results_under_different_filtering_threshold}
  \vspace{-0.22in}
\end{table*}

\paragraph{The values of the marker confidence are highly concentrated.} Since models are expected to express a wide range of confidence including extreme values which is necessary in mission-critical senarios \cite{human-confidence-extreme-value, human-confidence-extreme-value-2}, We expect the models to clearly differentiate the epistemic markers by obtaining a relatively uniform distribution and containing markers with a confidence near 0\% or 100\%. 
However, we found the I-AvgCV values typically range from approximately 0.14 and 0.24, demonstrating a concentrated distribution with minor difference. 
Additionally, only 4 out of 49 settings (dataset, model pair) include markers with confidence under 10\% when only those occur no less than 10 times are counted, indicating significant failure in expressing uncertainty. 

\subsection{Correlation between Performance and Marker Consistency}
\label{sec:model-capability-impact}

In Section~\ref{subsec:findings}, we notice that the statistics in \autoref{tab:combined_metrics} suggest that larger models demonstrate a better understanding of epistemic markers, as evidenced by lower C-AvgCV values and higher MRC values. 
In this section, we quantitatively evaluate the relationship between a model's accuracy and its corresponding C-AvgCV and MRC values to gain a deeper insight into the relationship of model ability and mastery of epistemic markers.

Specifically, for a given model \( M_k \), we use its average accuracy across all datasets as a comprehensive measure of its performance. 
We then compute the Pearson Correlation Coefficient between each model's overall accuracy and both its C-AvgCV and MRC.
The results show a correlation coefficient of \( -0.88 \) between model accuracy and C-AvgCV, and a correlation coefficient of \( 0.75 \) between model accuracy and MRC. 
These findings indicate a strong negative relationship between model accuracy and C-AvgCV and a strong positive relationship between model accuracy and MRC. 
This suggests that more powerful models exhibit greater stability in marker confidence across datasets, as well as a more consistent ordering of markers.

\subsection{Conclusion Robustness under Different Filtering Thresholds}

Our primary conclusions regarding marker consistency and model understanding of epistemic markers remain robust across various filtering thresholds.
As detailed in Table~\ref{tab:results_under_different_filtering_threshold}, even when increasing the filtering threshold to 50 or 100 occurrences, the observed trends persist: 
1) The C-AvgCV and MRC values consistently remain low, while the I-AvgCV values remain high. 
This pattern continues to suggest that models struggl to maintain consistent confidence across different datasets and to differentiate effectively between markers. 
2) The MAC values consistently remain high. 
This reinforces the strong positive relationship between model performance and its marker usage ability. 
Moreover, \textbf{the observed shortcomings extend even to frequently occurring markers} (with over 100 instances), indicating a severe challenge in the model's ability to reliably utilize epistemic expressions. 
This pervasive issue further substantiates our claims regarding the difficulties in ensuring reliable marker usage.

\subsection{Discussions}

This section aims to explain the observed differential in the consistent deployment of epistemic markers within in-domain versus out-of-domain contexts. Furthermore, we aim to investigate the distributional characteristics that potentially modulate the confidence associated with these markers.

Within similar distributions, models tend to maintain a stable ``preference'' for employing these markers to express uncertainty. 
For instance, in in-domain scenarios, responses on the test set leverage a consistent pattern of epistemic marker ``preference'' due to the similarity of the distribution. 
However, this consistency breaks down when the data distribution shifts. As the distribution changes, the model's preference usage for epistemic markers also changes, hindering the transfer of marker-based confidence to test sets from different datasets.

To further investigate this phenomenon, we conducted an in-depth analysis by calculating the average ECE values for each dataset to quantify the generalizability of the marker consistency from the training set to the test set. 
This involved averaging the ECE values across seven models in the in-domain scenario. 
Our analysis revealed that \textbf{datasets with greater component diversity generally exhibit a higher ECE}. 
This suggests that more complex or multi-sourced distributions make it challenging to transfer marker confidence from the training to the test set, as exemplified by datasets like MMLU and StrategyQA, which are not domain-specific. 
This observation further supports our claim that models perform well in in-domain scenarios primarily due to the consistency of the underlying data distribution. The details about the experiments are reported in Appendix~\ref{app:ablation-distribution}.






\section{Conclusion}


Our study evaluates whether LLMs can reliably express confidence using epistemic markers. 
We define marker confidence as the observed accuracy of responses containing specific markers, conduct extensive experiments and evaluate the results with several metrics.
The results show that the marker confidence shifts significantly under distribution changes, following the trend of model accuracy, which highlights poor stability. 
Additionally, models struggle to effectively differentiate between markers and maintain consistent marker rankings across datasets.
These findings suggest that the LLM generated markers to express their confidence is unreliable and requires improved alignment between verbal confidence and actual performance. 
Our work contributes to more consistent confidence estimation frameworks, ultimately facilitating reliable and trustworthy LLM responses.

\clearpage

\section*{Limitation}
Human language is remarkable for its complexity, variability, and rich connotations, particularly when expressing uncertainty.
Within a complete linguistic system, factors like sentence structure can significantly influence confidence, which is often difficult to quantify with epistemic markers alone. 
Moreover, in the context of long-form communication, it is clear that confidence cannot be simply measured by the confidence values of epistemic markers. 
To facilitate simplicity in evaluation and to focus on the study of epistemic markers, we adopt a relatively idealized approach: using epistemic markers generated by LLMs in closed-source QA tasks to represent the confidence of the responses while keeping them relatively brief. 
Additionally, epistemic markers may carry different meanings across cultures and languages. However, we only consider epistemic markers in English.

Despite our idealized conditions and using state-of-the-art models, LLMs still fail to consistently align epistemic markers with their true confidence levels, revealing that the issue lies not only with our approach but also with the models themselves. 
While they perform well in question-answering tasks, they do not truly understand epistemic markers \cite{zhou2023navigating}, struggle to express consistent confidence in these markers, and have difficulty aligning their confidence expressions with human expectations \cite{PerceptionOfLinguisticUncertainty}. This points to a deeper challenge in model behavior, suggesting that future research should focus on addressing  fundamental gaps in model linguistic alignment.


\section*{Ethics Statement}
Our work contains no offensive contents and have no potential risks.

\noindent\textbf{Licenses.}
We will share our code under the MIT license, allowing other researchers free access to our resources for research purposes. 
The datasets used in this paper, including BoolQ \cite{BoolQ}, StrategyQA \cite{StrategyQA}, GSM8K \cite{GSM8K}, MMLU \cite{MMLU}, CSQA \cite{CSQA}, MedMCQA \cite{MedMCQA}, and CaseHOLD \cite{CaseHOLD}, are shared under either the CC BY-SA license, Apache License Version 2.0, or the MIT License, all of which permit their use for research purposes.
As for language models, we access all open-source LMs via the Huggingface Hub~\cite{huggingface}. Our use of \textit{GPT-4o} \cite{GPT4o} and \textit{GPT-4o-mini} \cite{GPT4omini} is conducted through OpenAI's official website\footnote{https://platform.openai.com/docs/api-reference/introduction}.
All associated licenses permit user access for research purposes, and we have agreed to adhere to all terms of use.


\section*{Acknowledgements}
We thank the anonymous reviewers and the area chair for their constructive comments. 
The authors of this paper were supported by the ITSP Platform Research Project (ITS/189/23FP) from ITC of Hong Kong, SAR, China, and the AoE (AoE/E-601/24-N), the RIF (R6021-20) and the GRF (16205322) from RGC of Hong Kong,SAR, China.

\bibliography{custom}

\newpage
\appendix
\begin{center}
    {\Large\textbf{Appendices}}
\end{center}

\section{Related Work}
\label{app:related-work}


\paragraph{Confidence Estimation in LLMs.}

Confidence estimation in LLMs refers to the process of assessing a model's confidence in its output. 
Previous research on this topic can be categorized into white-box and black-box methods, distinguished by whether they utilize the model's internal information. 
White-box methods leverage the internal states of LLMs, with key approaches including information-based methods that analyze these inner states ~\cite{BenchmarkingUncertaintyQuantificationMethods, confidenceestimationsurvey, white=box-confidence-1}, such as perplexity ~\cite{perplexity, comparisonqa,zong2025critical}, the negative log probability of generated tokens ~\cite{TokenSAR}, and others. 
In the field of black-box methods, \citet{teachingmodelstoexpressuncertaintyinwords} first introduces the concept \textit{verbal confidence} that prompts LLM to output its confidence directly. 
Most subsequent methods are based on either directly prompting the model to generate an output or consistency sampling \cite{xiongcan, BSDetector, liu2024gprooft}. However, previous methods primarily focus on processing numerical values to estimate the LLM's confidence, leading to a research gap in exploring LLM confidence expression through linguistic patterns, especially epistemic markers.

\paragraph{Studies on Epistemic Markers.}
Epistemic markers are essential for expressing confidence in conversation, playing a key role in human-LLM interactions \cite{uncertaintyinNLP,guo2025mathematical}. Recent studies have examined the reliability of LLMs in mastering epistemic markers, primarily investigating their interpretation of various uncertainty expressions.
For instance, \citet{tangzhishengnaive} and \citet{PerceptionOfLinguisticUncertainty} use sentence templates with uncertainty expressions to prompt the model to assess overall confidence, though this approach is limited by the fixed nature of the templates, restricting generalizability. 
\citet{zhou2023navigating} and \citet{zhou2024relying} argue that LLMs often mimic marker distributions from training data rather than truly understanding them, with the latter highlighting a tendency for overconfidence.
\citet{robust-marker-intepret-bias} also examine epistemic markers but focus on robustness and biases in model interpretations. 
While these studies investigate LLM generation of epistemic markers, our work aligns most closely with \citet{intrinsicuncertaintyinwords}, which directly challenges the ability of LLMs to accurately convey confidence using epistemic markers.
However, they employ LLM-as-a-judge and few-shot prompting to assess the numerical confidence of uncertainty expressions, introducing potential bias \cite{llm-as-a-judge-bias1, llm-as-a-judge-bias2}.
Additionally, they use human judges to assess the quality of LLM judges, which essentially aligns the model's interpretation of markers with human understanding. This approach is similar to other works in the field, which also focus on aligning human and LLM recognition of epistemic markers.
However, as long as LLMs maintain a consistent framework for incorporating epistemic markers to express confidence, we can learn from its intrinsic mapping of marker usage and align it with human expectations through external means.

\section{Technical Details}

\subsection{Experiment Setup}
\label{app:experiment-setup}

\paragraph{\textbf{Models}}
We incorporate a range of commonly used LLMs of different scales, including Llama-3.1-8B-Instruct \cite{Llama}, Qwen2.5-7B-Instruct, Qwen2.5-14B-Instruct, Qwen2.5-32B-Instruct \cite{Qwen}, Mistral-7B-Instruct-v0.3 \cite{Mistral}, GPT-4o \cite{GPT4o}, GPT-4o-mini \cite{GPT4omini}. For all models, we use a temperature of 0.5 to balance between logical consistency and creativity. All the open-source models are run on 4 NVIDIA A6000 (40G) GPUs with BF16.

\begin{figure}[t]
\centering
    \begin{minipage}{0.5\textwidth}
        \centering
        \includegraphics[width=\textwidth]{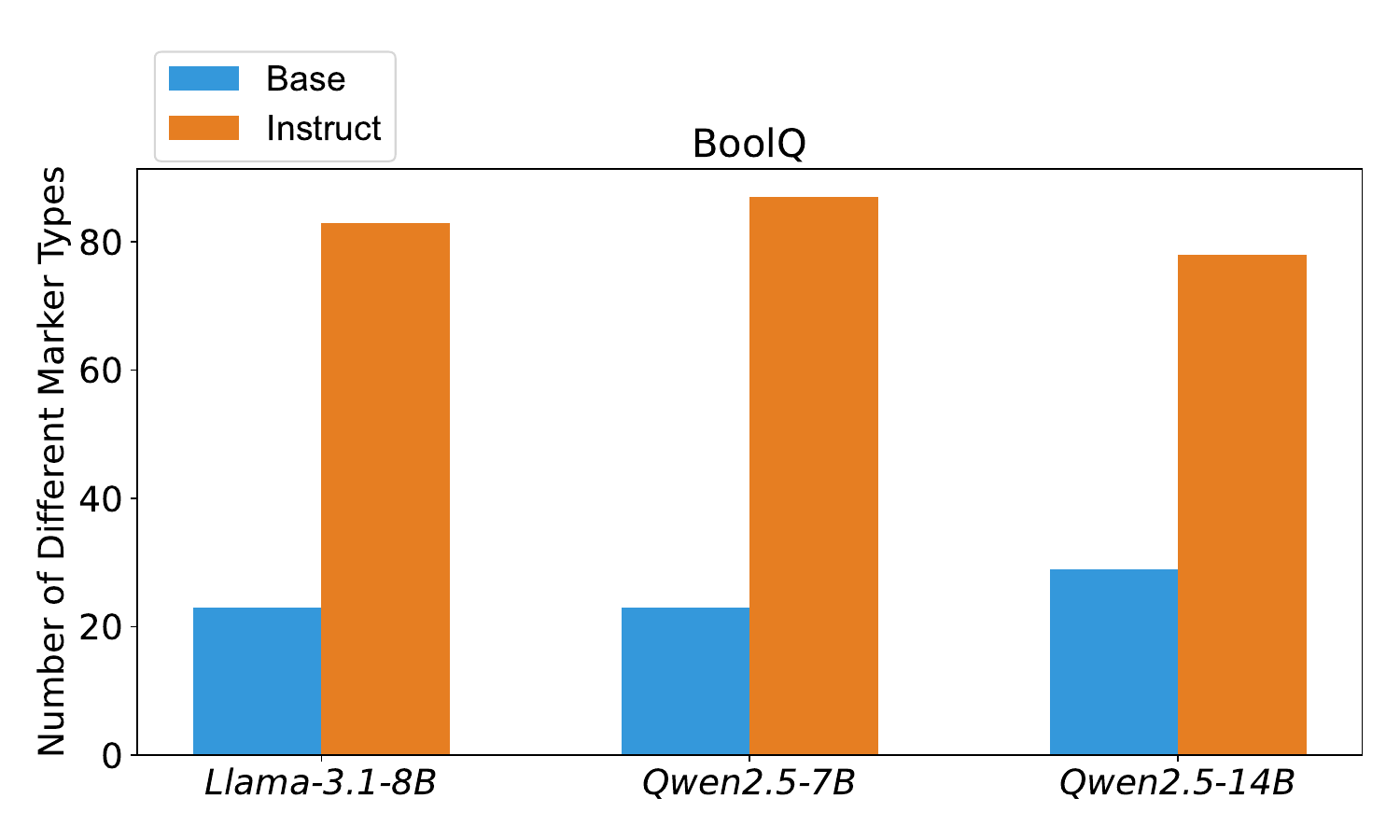}
    \end{minipage}%
    \hfill
    \begin{minipage}{0.5\textwidth}
        \centering
        \includegraphics[width=\textwidth]{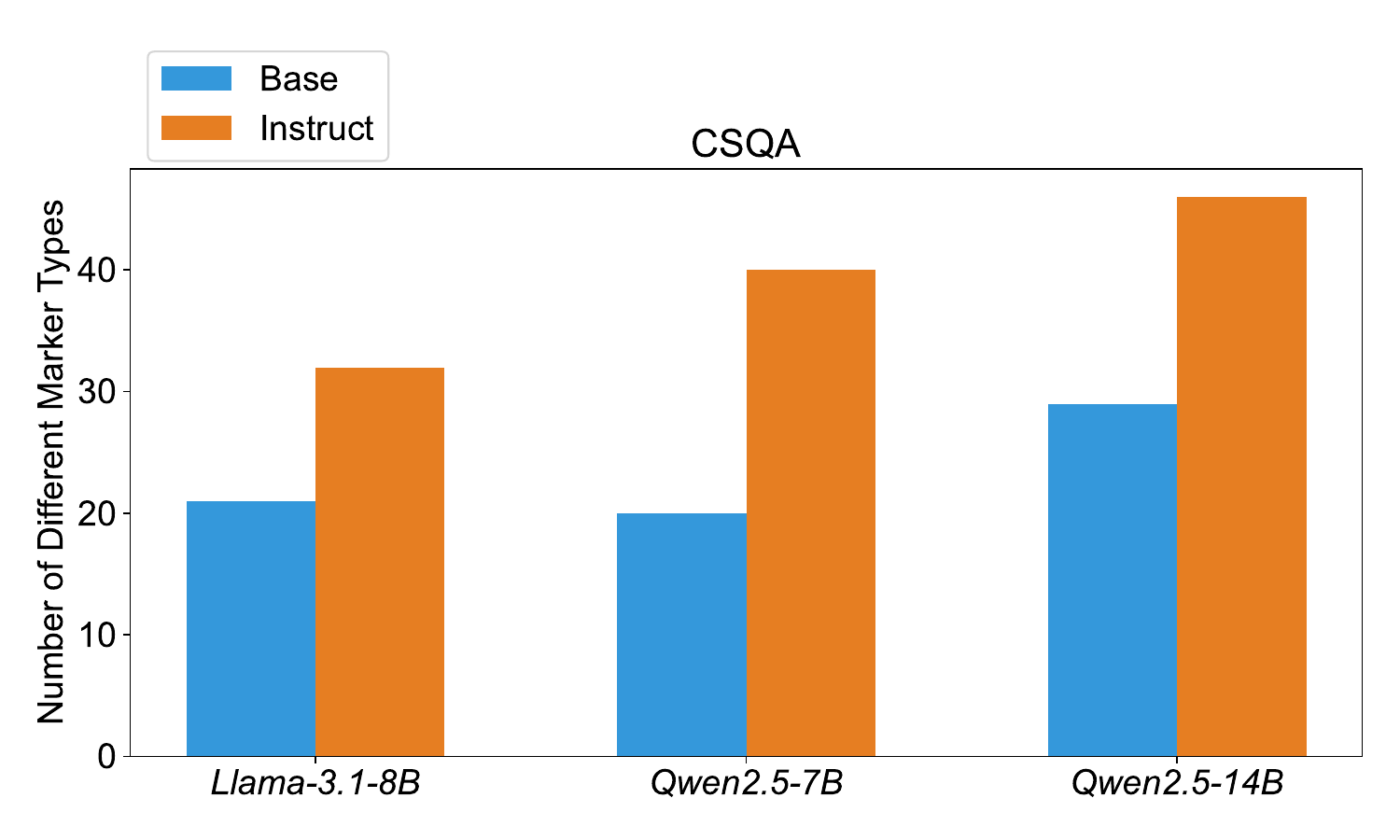}
    \end{minipage}
    \caption{The number of epistemic markers that six different models generated in BoolQ and CSQA dataset. The results indicate that the instruct-tuned models exhibit much better linguistic diversity than base models in expressing confidence, which is desired by our experiment.}
    \label{fig:base-instruct-models}
\end{figure}

We observed that instruction-tuned models exhibit greater variation and demonstrate better linguistic diversity when using epistemic markers.
Although we also tested some base models, we found that for the same dataset, they emitted significantly fewer markers compared to the instruction-tuned models (See \autoref{fig:base-instruct-models}) .
As a result, we chose to focus our experiments on instruction-tuned models instead.

\paragraph{\textbf{Datasets}} We benchmark the LLMs' responses with confidence expression using epistemic markers using the following seven datasets requiring knowledge in different domains: 
1) Factual and commonsense knowledge: BoolQ \cite{BoolQ}, StrategyQA \cite{StrategyQA}, CSQA \cite{CSQA}. 
2) Mathematical reasoning: GSM8K \cite{GSM8K}.
3) Medical reasoning: MedMCQA \cite{MedMCQA}. 
4) Law reasoning: CaseHOLD \cite{CaseHOLD}. 
5) Mixed factual datasets: MMLU \cite{MMLU}. 

\paragraph{\textbf{Prompts}}
\label{app:prompt}

The prompts used in our experiments are shown in \autoref{tab:prompt}. We performed five permutations on the prompt with \textit{Qwen2.5-7B-Instruct} and \textit{Mistral-7B-Instruct-v0.3} on BoolQ, StrategyQA, CSQA, MedMCQA and found that both accuracy and C-AvgCV, MAC, MRC and I-AvgCV values varied only slightly. Consequently, we randomly selected one version and conducted all subsequent experiments using it.

\subsection{Model Sources}

This section clarifies the sources of the models used in our study. For methods involving LLMs, we utilize their instruction fine-tuned versions (see Appendix~\ref{app:experiment-setup} for more details) accessed via the Hugging Face Hub \cite{huggingface}. Specifically, for \textit{Llama-3.1-8B-Instruct}, we employ the version \textit{meta-llama/Llama-3.1-8B-Instruct}. The models related to Qwen include \textit{Qwen/Qwen2.5-7B-Instruct}, \textit{Qwen/Qwen2.5-14B-Instruct}, and \textit{Qwen/Qwen2.5-32B-Instruct}. For \textit{Mistral-7B-Instruct-v0.3}, we use \textit{mistralai/Mistral-7B-Instruct-v0.3}.

\subsection{Detailed Implementation}

\paragraph{\textbf{Data Preprocessing}}

We applied data preprocessing methods to the datasets. For GSM8K, we transformed it into a binary-question dataset for convenience. Specifically, for half of the questions \( Q_i \) with even index in GSM8K, we first extracted the correct answer \( A_i \), then used a question template to create a binary question by incorporating \( (Q_i, A_i) \) and setting the correct binary answer to ``yes.'' For the remaining half of the questions \( Q_j \), we randomly selected an answer \( A_j \) different from the correct answer, and used the same question template to create a binary question by incorporating \( (Q_j, A_j) \) and setting the correct binary answer to ``no.'' The question template and two examples are given in \autoref{tab:GSM8K-preprocessing}. For the training set of the MMLU, due to its massive size, we randomly sampled a subset of 20000 QA-pairs for the MMLU training set. For MedMCQA dataset, for the convenience of evaluation, we pick the subset of the answer with only one correct answer. We then randomly sample 9686 QA-pairs for the MedMCQA training set and 2422 for its test set. For BoolQ, we did not expose the model to the ``passage'' part and treated it as a closed-book question-answering dataset in our experiment. Since CaseHOLD isn't explicit split into training set and test set, we divide the former 80\% as training set and the rest as test set. A detailed statistics for our dataset usage is on \autoref{tab:dataset-usage}.

\begin{table}[h]
    \centering
    \begin{tabular}{lcc}
        \toprule
        Dataset     & Train Size & Test Size \\ \midrule
        BoolQ       & \phantom{0}9427       & \phantom{0}3270      \\
        StrategyQA  & \phantom{0}2061       & \phantom{0}\phantom{0}229       \\
        GSM8K       & \phantom{0}7473       & \phantom{0}1319      \\
        MMLU        & 20000      & 14041     \\
        CSQA        & \phantom{0}9741       & \phantom{0}1221      \\
        MedMCQA     & \phantom{0}9686       & \phantom{0}2422      \\
        CaseHOLD    & \phantom{0}8396       & \phantom{0}2099      \\ \bottomrule
    \end{tabular}
    \caption{A detailed statistcs of our dataset usage.}
    \vspace{-0.2in}
\label{tab:dataset-usage}
\end{table}

\begin{table*}[h]
\centering
\begin{tabular}{p{0.2\textwidth}|p{0.35\textwidth}|p{0.35\textwidth}}
\toprule
 & \textbf{Binary Question} & \textbf{Multiple Choice Question} \\
\midrule
\textbf{Eliciting epistemic markers} &
\textit{User}: The following is a binary question. When responding, answer with a binary answer from \textbf{[choices]} and incorporate only one epistemic marker to reflect your confidence level. You must include your binary answer at the beginning of your response then respond with the epistemic markers in a concise and brief manner. \newline The question is: \textbf{[Question]} \newline And your answer is: & 
\textit{User}: The following is a multiple choice question. When responding, answer with a letter from \textbf{[choices]} and incorporate only one epistemic marker to reflect your confidence level. You must include your choice of letter at the beginning of your response then respond with the epistemic markers in a concise and brief manner. \newline The question is:  \textbf{[Question]} \newline The options are:   \textbf{[Options]} \newline And your answer is: \\
\midrule
\textbf{Eliciting numerical values} & 
\textit{User}: The following is a binary question. When responding, answer with a binary answer from \textbf{[choices]} and incorporate a number between 0 and 100 to reflect your confidence level. You must include your binary answer at the beginning of your response then respond with the confidence score in a concise and brief manner. \newline The question is: \textbf{[Question]} \newline
And your answer is: & 
\textit{User}: The following is a multiple choice question. When responding, answer with a letter from \textbf{[choices]} and incorporate a number between 0 and 100 to reflect your confidence level. You must include your choice of letter at the beginning of your response then respond with the confidence score in a concise and brief manner. \newline The question is: \textbf{[Question]} \newline The options are: \textbf{[Options]} \newline And your answer is: \\
\bottomrule
\end{tabular}
\caption{Our prompt to elicit epistemic markers and numerical confidence values. The text inside the square brackets is filled with actual content in the dataset. Specifically, choices are capital letters that represent the options (e.g., ``A, B, C, D'' or ``A, B, C, D, E'') for multiple choice questions and ``yes or no'' for binary questions.}
\label{tab:prompt}
\end{table*}

\begin{center}
\begin{table*}[h]{
\begin{tabular}{p{0.97\textwidth}}
\toprule
\textbf{Question Template} \\
For the question \textbf{[original question]}, is the answer \textbf{[original answer]} its correct answer? \\ \hline 
\textbf{Sample 1} \\ $Q_i$: Each bird eats 12 beetles per day, each snake eats 3 birds per day, and each jaguar eats 5 snakes per day. If there are 6 jaguars in a forest, how many beetles are eaten each day? \\\\
$A_i$: 1080 (The correct answer is 1080)\\\\
Integrated binary question: For the question `Each bird eats 12 beetles per day, each snake eats 3 birds per day, and each jaguar eats 5 snakes per day. If there are 6 jaguars in a forest, how many beetles are eaten each day?'', is the answer 1080 its correct answer?\\ \\
Binary answer: Yes. \\
\hline
\textbf{Sample 2}\\
$Q_j$: James writes a 3 - page letter to 2 different friends twice a week. How many pages does he write a year?\\ \\
$A_j$: 223 (Randomly generated answer, the correct answer is 624)\\ \\
Integrated binary question: For the question `James writes a 3 - page letter to 2 different friends twice a week. How many pages does he write a year?'', is the answer 223 its correct answer?\\ \\
Binary answer: No. \\
\bottomrule
\end{tabular}}
\caption{Data pre-processing method used in GSM8K. The text inside the square brackets is replaced by actual content in the dataset. Sample 1 keeps the original correct answer and incorporate it into the binary answer while setting the binary answer to ``Yes.'' Sample 2 randomly generates a wrong answer and set the binary answer to ``No.''}
\label{tab:GSM8K-preprocessing}
\end{table*}
\end{center}

\paragraph{Epistemic Marker Extraction}

For each model, we extract the epistemic marker from each response by few-shot prompting \cite{few-shot-prompting,wang2025prospect,wang2025diversity} the same model to recognize the epistemic markers emitted by itself. 
We manually examined a subset of each dataset and find out most of them are able to recognize the epistemic markers. 
For the unrecognized ones or the one that didn't match the desired format, we uniformly use \textit{GPT-4o-mini} to extract its epistemic markers. 
According to \citet{zhou2024relying}, models are relunctant to express confidence in words, so we also provided responses that did not include any epistemic markers as few-shot samples, and those with no markers are grouped together as a special epistemic marker.

\subsection{Evaluation Metrics}
This section introduces the evaluation metrics in detail and explain our experiment settings.

\subsubsection{Detailed Implementation and Formulas}
\label{app:evaluation-metrics-appendix}

Our evaluation metrics are categorized into three kinds: ECE-based \cite{xiongcan}, CV-based \cite{cv-value} and Spearman/Pearson coefficient based \cite{Spearman, Pearson}, aiming to reflect the calibration error, the degree of dispersion and correlation respectively.

\paragraph{\textbf{NumECE, I-AvgECE, and C-AvgECE}}

To evaluate the calibration of model-generated confidence, we introduce three metrics: NumECE, I-AvgECE, and C-AvgECE. Since the latter two are based on generalization of marker confidence, these metrics assess the model's stability on marker confidence, considering both within-domain (I-AvgECE) and cross-domain (C-AvgECE) scenarios and comparing with number-based methods.

\textbf{NumECE} measures the overall calibration of the model’s outputted numerical confidence. For each dataset \(D_j\), we compute the expected calibration error (ECE-num) based on the model’s numerical confidence on the test set. The final NumECE is the average of these ECE values across all datasets, providing a comprehensive evaluation of the model’s confidence calibration.

\textbf{I-AvgECE} focuses on the model’s performance within the same domain, where the training and test datasets are identical. For each dataset \(D_p\), we calculate the marker-based expected calibration error (ECE-mar) by using the marker's confidence \(\textit{Conf}(W_i, D_p, M_k)\) obtained from the training set of  \(D_p\) on the test set of it when the model also emits $W_i$ as confidence expression. The final I-AvgECE is obtained by averaging these values across all datasets as well.

\textbf{C-AvgECE} evaluates the model's ability to generalize its marker confidence across different datasets. For each pair of distinct datasets \( (D_p, D_q) \), where \( D_p \neq D_q \), we calculate the marker-based expected calibration error \(\text{ECE-mar}(D_p, D_q, M_k)\). This is done by using the model's confidence \(\textit{Conf}(W_i, D_p, M_k)\) on the training dataset \(D_p\) to estimate the model’s marker confidence on the test set of \(D_q\), thereby measuring the model's ability to transfer its marker confidence to a new dataset. The final C-AvgECE is computed by averaging the ECE-mar values across all dataset pairs, providing insight into the model's robustness in handling cross-distribution variations.
The mathematical formulas of three ECE-based metrics is given by: 



{\small
\begin{align*}
\text{NumECE} = \frac{1}{|D|} \sum_{D_j \in D} \text{ECE-num}(D_j, M_k),
\end{align*}}

{\small
\begin{align*}
\text{I-AvgECE} = \frac{1}{|D|} \sum_{D_p \in D} \text{ECE-mar}(D_p, D_p, M_k),
\end{align*}}

{\small
\begin{align*}
\text{C-AvgECE}  = \frac{\sum_{\substack{D_p, D_q \in D \\ D_p \neq D_q}} \text{ECE-mar}(D_p, D_q, M_k)}{|D| * (|D| - 1)}  
\end{align*}}

\noindent where \( |D| \) is the total number of datasets (\( D_1, D_2, \dots, D_n \)). Note that for all ECE values, we use a ECE bin number of \textit{N}, where \textit{N} is the number of confidence predictions.

\paragraph{\textbf{I-AvgCV and C-AvgCV}}

To quantify the concentration and variation of marker confidence, we propose I-AvgCV and C-AvgCV. These metrics assess how dispersed and consistent marker confidence is within individual datasets and across datasets, respectively.

\textbf{I-AvgCV} measures the concentration of marker confidence within a single dataset. For each dataset \(D_j\), we calculate the coefficient of variation (CV) of the confidence of different markers \( \textit{Conf}(W_i, D_j, M_k) \). The final I-AvgCV is the average CV value across all datasets, providing an overall measure of confidence concentration.

It is important to note that while we expect the distribution of marker confidence to be more dispersed, we are not claiming that greater dispersion is inherently better. 
Our desired result for models is to cover a relatively wide range of confidence values across all markers, while also clearly differentiating between different markers. This would facilitate more effective confidence expression in a variety of scenarios. However, as shown in \autoref{tab:combined_metrics}, the average I-AvgCV value marker is lower than 0.2, which indicates that the marker confidence is highly concentrated, leading us to conclude that the model fails to clearly differentiate between the markers.

\textbf{C-AvgCV} evaluates the consistency of marker confidence across datasets. For each shared marker (markers that appear in each dataset) \(W_i\), we compute the CV of the marker confidence across different datasets and then average these values over all shared markers. The final C-AvgCV quantifies the consistency of model-generated confidence across multiple datasets.

The mathematical formulations for \(\text{I-AvgCV}\) and \(\text{C-AvgCV}\) are given by:



{\small
\begin{align*}
CV(D_j, M_k) = \frac{\sigma(D_j, M_k)}{\mu(D_j, M_k)},
\end{align*}}

{\small
\begin{align*}
\text{I-AvgCV}(M_k) = \frac{1}{|D|} \sum_{j=1}^{|D|} CV(D_j, M_k),
\end{align*}}

{\small
\begin{align*}
CV(W_i, M_k) = \frac{\sigma(W_i, M_k)}{\mu(W_i, M_k)},
\end{align*}}

{\small
\begin{align*}
\text{C-AvgCV}(M_k) = \frac{1}{|W|} \sum_{i=1}^{|W|} CV(W_i, M_k)
\end{align*}}

\noindent where |$W$| is the number of shared markers across every datasets for model $M_k$, \( \sigma(D_j, M_k) \) is the standard deviation of the confidence scores for all markers in dataset \( D_j \) for model \( M_k \), \( \mu(D_j, M_k) \) is the mean of the confidence scores for all markers in dataset \( D_j \) for model \( M_k \), \( \sigma(W_i, M_k) \) is the standard deviation of the confidence scores for the marker \( W_i \) across different datasets for model \( M_k \) and \( \mu(W_i, M_k) \) is the mean of the confidence scores for the marker \( W_i \) across different datasets for model \( M_k \).

\paragraph{MRC}




To assess the alignment of marker rankings across datasets, we introduce a metric based on the Spearman rank correlation coefficient: \textit{Marker Rank Correlation} (MRC). This metric capture the degree of consistency in marker confidence rankings across datasets.

Specifically, For each pair of datasets \( (D_p, D_q) \), we compute the Spearman rank correlation coefficient between the confidence rankings of shared markers. The final MRC for the model is the average correlation across all dataset pairs.
The mathematical formulations for MRC are given by:



{\small
\begin{align*}
\text{MRC} = \frac{1}{|P|} \sum_{\substack{(D_p, D_q) \in P \\ D_p \neq D_q}} \rho(D_p, D_q)
\end{align*}}

\noindent where $W_1, \dots, W_i$ are all the epistemic markers that shared by $D_p$ and $D_q$, \( \text{S}(X, Y) \) denotes the Spearman rank correlation coefficient between the rankings of \( X \) and \( Y \) and \( \text{Conf}(W_i, D_j, M_k) \) represents the confidence of marker \( W_i \) for model \( M_k \) on dataset \( D_j \).



\textbf{MAC}
To analyze whether the confidence of markers and the accuracy of the model are positively correlated, we propose the \textit{Marker Accuracy Correlation} (MAC) based on the Pearson correlation coefficient. 

Specifically, for a given model \( M_k \), we consider the confidence of a specific shared marker \( W_i \), which is present across all datasets associated with \( M_k \). 
We then compute the Pearson correlation coefficient between the set of marker confidences across these datasets and the model's overall accuracies on the same datasets. 
Finally, we compute the average of the correlation coefficients \( \rho(W_i, M_k) \) across all shared markers \( W_i \) to obtain the overall correlation coefficient for the model, denoted as \( \text{MAC}(M_k) \). It's mathematical formula is given by:

{\small
\begin{align*}
\text{MAC}(M_k) = \frac{1}{|W|} \sum_{W_i \in W} \rho(W_i, M_k),
\end{align*}}

\noindent where \( W \) is the set of all shared markers \( W_i \), |$W$| is the number of all shared markers and \( \rho(W_i, M_k) \) is the Pearson correlation coefficient between the confidence of marker \( W_i \) and the model's accuracy on all the datasets.

These metrics provide a quantitative assessment of the consistency and concentration of model-generated confidence values across different datasets.

\begin{center}
   \begin{table}[h!] 
     \centering 
     \begin{tabular}{c c}
       \toprule
       \textbf{Threshold} & \textbf{Confidence Interval} \\
       \hline
       10 & $\sim$23\% \\
       50 & $\sim$10\% \\
       200 & \phantom{0}$\sim$5\% \\
       \bottomrule
     \end{tabular}
     \caption{The exact confidence interval of different filtering thresholds.} 
     \label{tab:confidence-interval} 
   \end{table}
   \vspace{-0.2in}
\end{center}

\subsubsection{Details on Marker Filtering Strategies}
\label{app:why-filtering}

The filtering is necessary because our method of quantifying marker confidence is based on accuracy, which is affected by the confidence interval. 
If the sample size for a particular marker is too small, its corresponding confidence values can be heavily influenced by random variations. 
For instance, if the marker ``unsure'' appears only once in the training set and the response happens to be correct, the marker confidence for ``unsure'' would be 100\%, which may not accurately reflect the model's true intent. 
Previous works also show that confidence expression could be more reflective for humans when the it is determined in a crowd-source manner \cite{confidence-is-accurate-in-crowd-source-scenarios(explaining-10)-and-people-intepret-differently-to-markers}, which supports our setting.

Furthermore, we have enough shared markers (more than ten for each model) after implementing the filtering, which ensures the reliability of our experiment.

On the other hand, all epistemic markers obtained from the training set are used for the experiment related to ECE values. 
Since the estimated confidence for each question in the test set is derived from the marker confidence in the training set, it is essential to ensure that the vast majority of markers in the test set can be mapped to corresponding markers in the training set. 
This approach is reasonable since the low frequency of marker occurrences results in minimal impact on the overall calibration performance, which ensures both completeness and reliability.

\section{Distribution Analysis}
\label{app:ablation-distribution}

Our investigation primarily focused on the influence of \textbf{distribution diversity} and \textbf{difficulty} on the generalizability of marker confidence. We measured this generalizability by calculating the Average ECE value across various models for each specific dataset.

Table~\ref{tab:diversity-inves} illustrates that multi-domain datasets, specifically StrategyQA (14.42\%) and MMLU (15.97\%), exhibit significantly higher Average ECE values compared to single-domain datasets such as GSM8K (Math, 6.68\%), MedMCQA (Medical, 6.25\%), and CaseHOLD (Law, 9.53\%). This finding leads to the conclusion that a more diverse data distribution impedes the transfer of models' marker confidence preferences from the training to the test set~\cite{liu2025costbench,liu2026naacl}.

\begin{center}
\small
   \begin{table}[h!]
     \centering
     \begin{tabular}{l c c}
       \toprule
       \textbf{Dataset} & \textbf{Domain} &\textbf{Average ECE} \\
       \hline
       StrategyQA & Multi-domain & 14.42 \\
       MMLU & Multi-domain & 15.97 \\
       GSM8K & Math & \phantom{0}6.68 \\
       MedMCQA & Medical & \phantom{0}6.25 \\
       CaseHOLD & Law & \phantom{0}9.53 \\
       \bottomrule
     \end{tabular}
     \caption{The Average ECE values across different models for certain dataset. We found that multi-domain datasets exhibit higher average ECE values than single-domain ones.}
     \label{tab:diversity-inves}
   \end{table}
\vspace{-0.35in}
\end{center}

Additionally, we observed that a substantial \textbf{difficulty gap} between the training and test sets of the marker confidence also compromises its transferability. 
As shown in Table~\ref{tab:difficulty-inves}, we utilized a challenging math dataset, MATH-500, and a simpler math dataset, GSM8K, to compute the MRC value, representing the relevance of marker rankings across these two datasets. 
For both \textit{Qwen2.5-7B-Instruct} (0.38) and \textit{Mistral-7B-Instruct-v0.3} (0.26), the models demonstrated low relevance in marker rankings obtained from datasets with differing difficulties. 
This indicates that the \textbf{difficulty of the data distribution} can significantly affect the generalizability of marker confidence preferences.

\begin{center}
\small
   \begin{table}[h!]
     \centering
     \begin{tabular}{l c}
       \toprule
       \textbf{Model} &\textbf{MRC} \\
       \hline
       \textit{Qwen2.5-7B-Instruct} & 0.38 \\
       \textit{Mistral-7B-Instruct-v0.3} & 0.26 \\
       \bottomrule
     \end{tabular}
     \caption{The MRC value for two models across MATH-500 and GSM8K. The results shows that marker rankings obtained from the two datasets are relevant to little extent.}
     \label{tab:difficulty-inves}
   \end{table}
\end{center}

\end{document}